\documentclass[preprint,12pt]{elsarticle}

\usepackage{caption}
\usepackage{graphicx}
\usepackage{amssymb}
\usepackage{algorithm}
\usepackage{algpseudocode}
\usepackage{caption}
\usepackage{subcaption}
\usepackage{multirow}
\usepackage{amsmath}
\usepackage{hyperref}
\usepackage{comment}
\usepackage{algorithmicx}
\usepackage{lineno}
\usepackage{xcolor}
\usepackage{booktabs}
\usepackage{eurosym}
\usepackage{xurl}

\makeatletter
\def\ps@pprintTitle{%
  \let\@oddhead\@empty
  \let\@evenhead\@empty
  \let\@oddfoot\@empty
  \let\@evenfoot\@empty
}
\makeatother

\begin{document}

\begin{frontmatter}



\title{Forecast Aware Deep Reinforcement Learning for Efficient Electricity Load Scheduling in Dairy Farms}


\author[1]{Nawazish Ali}
\author[2]{Rachael Shaw}
\author[1]{Karl Mason}

\affiliation[1]{organization={School of Computer Science, University of Galway},
    addressline={},
    city={Galway},
    postcode={H91 TK33},
    state={},
    country={Ireland}}

\affiliation[2]{organization={School of Computing, Atlantic Technological University},
    addressline={},
    city={Galway},
    postcode={H91 T8NW},
    state={},
    country={Ireland}}


\begin{abstract}

Dairy farming is an energy intensive sector that relies heavily on grid electricity. With increasing renewable energy integration, sustainable energy management has become essential for reducing grid dependence and supporting the United Nations Sustainable Development Goal 7 on affordable and clean energy. However, the intermittent nature of renewables poses challenges in balancing supply and demand in real time. Intelligent load scheduling is therefore crucial to minimize operational costs while maintaining reliability. Reinforcement Learning has shown promise in improving energy efficiency and reducing costs; however, most RL-based scheduling methods assume complete knowledge of future prices or generation, which is unrealistic in dynamic environments. Moreover, standard PPO variants rely on fixed clipping or KL-divergence thresholds, often leading to unstable training under variable tariffs. To address these challenges, this study proposes a Deep Reinforcement Learning framework for efficient load scheduling in dairy farms, focusing on battery storage and water heating under realistic operational constraints. The proposed Forecast-Aware PPO incorporates short-term forecasts of demand and renewable generation using hour-of-day and month-based residual calibration, while the PID-KL PPO variant employs a proportional–integral–derivative controller to regulate KL-divergence for stable policy updates adaptively. Trained on real-world dairy farm data, the method achieves up to 1\% lower electricity cost than PPO, 4.8\% than DQN, and 1.5\% than SAC. For battery scheduling, PPO reduces grid imports by 13.1\%, demonstrating scalability and effectiveness for sustainable energy management in modern dairy farming.

\end{abstract}




\end{frontmatter}



\section{Introduction}
The increasing global population has driven a substantial rise in the demand for dairy products, reinforcing the importance of dairy farming within the agricultural sector \cite{dairy_growth}. Projections from the OECD-FAO Agricultural Outlook 2020–2029 indicate that global milk production is expected to grow at an annual rate of 1.6\%, reaching approximately 997 million metric tons by 2029. This rising demand has led to increased milk production and an expansion in the global export of dairy products \cite{dairy_export}. 

Dairy farms are inherently energy-intensive, requiring substantial electricity for various operations, including milking, cooling, and storage \cite{dairy_operations}. The growing scale of milk production further amplifies electricity consumption, making energy efficiency a crucial concern for the industry. Recent research by Dew et al. in 2020 has identified several energy-intensive tasks that can be deferred during peak electricity demand hours to reduce peak load demand e.g, water heater, water pumps, effluent pump, and milk chillers \cite{dew2021reducing}. Given that dairy farms frequently import electricity from the national grid, which primarily relies on fossil fuel-based power generation, minimizing this dependency directly supports carbon emission reductions and fosters a more sustainable energy framework \cite{grid_gen}. Therefore, innovative approaches to manage the energy-intensive processes involved in dairy farming are essential for ensuring sustainability.

With increasing demand for electricity, there has been a significant increase in the integration of renewable energy sources for sustainability in dairy farming \cite{ahdb_renewable_energy}. The adoption of renewable energy shows the industry's focus on reducing carbon footprints and adopting sustainable energy sources. However, the inherent intermittency of renewable energy generation poses a significant challenge, as fluctuations in energy supply can disrupt farm operations \cite{intermitency}. This variability emphasizes the need for efficient energy management techniques to reduce the effect of variations between energy generation and consumption.

Effective battery scheduling plays a crucial role in mitigating the intermittent nature of renewable energy sources by ensuring efficient energy storage and utilisation. By strategically charging and discharging batteries based on real-time electricity prices, renewable energy availability, and on-farm electricity demand, dairy farms can enhance energy efficiency while reducing reliance on grid imports. Similarly, water heater scheduling is crucial due to its significant energy consumption; inefficient scheduling of water heaters can lead to excessive energy costs, increased peak demand, and suboptimal utilisation of renewable energy.

Recent advancements in Artificial Intelligence (AI), particularly Deep Reinforcement Learning (DRL) \cite{ai_advancement}, provide a promising solution to these challenges by facilitating the integration of renewable energy sources and the effective management of battery storage in dairy farming. DRL is particularly well-suited for decision-making in complex and uncertain environments, as it learns optimal actions through continuous trial and error \cite{deepcomplextask}. This capability makes DRL ideal for optimizing energy usage and storage in scenarios with fluctuating renewable energy availability. By leveraging DRL algorithms, dairy farms can dynamically adjust their energy consumption and storage based on real-time data, maximizing the utilisation of renewable energy while enhancing overall operational efficiency.


The main objective of this study is to investigate  Reinforcement Learning for load scheduling in dairy farms. Several RL algorithms are implemented and compared within a dairy farm load scheduling environment to evaluate their performance under realistic operational constraints. The proposed framework extends the standard Proximal Policy Optimization (PPO) \cite{ppo} by incorporating forecasting-aware Aware Proximal Policy Optimization(F-PPO) based learning, enabling the agent to anticipate variations in renewable generation and electricity prices through short-term forecasting signals. Furthermore, a Proportional Integral Derivative Kullback Leibler (PID-KL) adaptive mechanism is integrated to dynamically regulate policy updates, enhancing training stability under fluctuating electricity prices.
To comprehensively evaluate the effectiveness of the proposed approach, we compare it with three reinforcement learning algorithms: PPO, Deep Q-Network (DQN) \cite{dqn}, and Soft Actor Critic(SAC) \cite{sac}. PPO serves as a strong on-policy benchmark for assessing stability and sample efficiency, while DQN provides a value-based contrast that highlights the challenges of discrete Q-learning in stochastic, constraint-driven environments. SAC, an off-policy actor–critic method with entropy regularization, enables comparison against a stochastic exploration baseline.  By combining DRL-based forecasting optimization methods, this study demonstrates a significant step toward sustainable and cost-efficient energy management in modern dairy farming systems.

The main contributions of this research are highlighted below:  
\begin{itemize}  
    \item Developed a forecasting-integrated PPO framework for water heater scheduling to anticipate renewable generation and electricity demand variations.
    \item Incorporated a PID-KL controller and embedded key operational constraints to ensure stable training and realistic scheduling under dynamic electricity pricing.
    \item To comprehensively evaluate the effectiveness of the proposed algorithm, we compare its performance with different RL algorithms.
    \item Implemented the PPO-based approach for battery scheduling, enabling efficient energy management.
\end{itemize} 

\section{Literature Review and Background}

\subsection{Reinforcement Learning (RL)}
RL is a crucial branch of AI that enables an agent to make optimal decisions through continuous interaction with its environment. In RL, the agent learns a policy $\pi$ aimed at maximizing cumulative rewards, which it obtains by performing actions and observing outcomes. The learning process involves iterative exploration, where the agent observes environmental states \textbf{\textit{S}}, chooses and executes actions \textbf{\textit{A}}, and receives rewards from the environment \textbf{\textit{R}}. The environment transitions between states according to a transition probability \textbf{\textit{P}}. The interaction is formally represented as a Markov Decision Process (MDP), characterized by the tuple (\textbf{\textit{S, A, P, R, $\gamma$}}), where $\gamma$ denotes the discount factor controlling the balance between immediate and future rewards. The agent optimizes a policy based on a value function, aiming to estimate actions that yield the highest expected cumulative reward. The policy, denoted as $\pi$, guides the agent's decisions to maximize long-term reward. This process is mathematically described by the state-action value function, as illustrated in Equation~\ref{bellman}:

\begin{equation}
Q^\pi(s,a) = \mathbb{E}\left[ R_{t+1} + \gamma Q^\pi(S_{t+1}, A_{t+1}) \mid S_t=s, A_t=a \right]
\label{bellman}
\end{equation}

Equation~\ref{bellman} defines the expected cumulative reward obtained by performing action $a$ in state $s$, considering both the immediate reward $R_{t+1}$ and discounted future rewards. The iterative training cycle and interaction between the RL agent and the environment are depicted clearly in Figure~\ref{Figure_graph_drl}.


\subsection{Deep Reinforcement Learning}

DRL extends traditional Reinforcement Learning (RL) by integrating deep neural networks to approximate value functions or policies, enabling decision-making in high-dimensional state and action spaces. Unlike conventional RL methods that rely on tabular representations, DRL employs function approximation to generalize learning across complex environments. The agent interacts with an environment, observes states \(\mathbf{S}\), selects actions \(\mathbf{A}\) based on a policy \(\pi\), and receives rewards \(\mathbf{R}\), leading to state transitions. This process is modeled as a Markov Decision Process (MDP) represented by the tuple \((\mathbf{S, A, P, R, \gamma})\), where \(\mathbf{P}\) denotes transition probabilities, and \(\gamma\) is the discount factor controlling the trade-off between immediate and future rewards.

In DRL, a deep neural network parameterized by \(\theta\) is used to approximate the action-value function \(Q(s, a; \theta)\), which estimates the expected cumulative reward. The optimization follows the Bellman equation with function approximation:

\begin{equation}
Q(s, a; \theta) = \mathbb{E} \left[ R_{t+1} + \gamma \max_{a'} Q(S_{t+1}, a'; \theta) \mid S_t = s, A_t = a \right]
\label{dqn}
\end{equation}

\begin{figure}[H]
\centering
\includegraphics[width=0.60\textwidth]{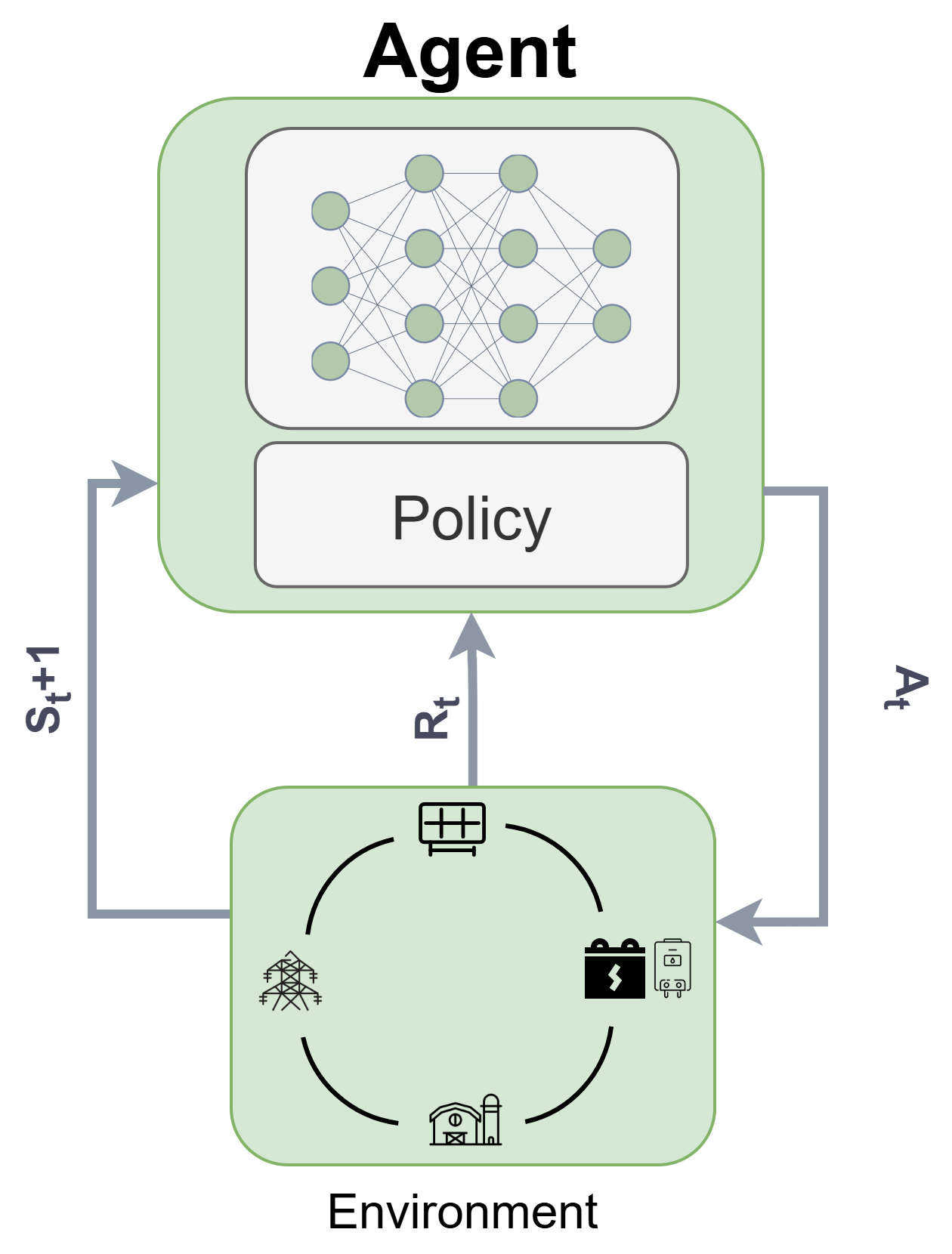}
\caption{The flow diagram of the Deep Reinforcement Learning process.} \label{Figure_graph_drl}
\end{figure}

Equation \ref{dqn} represents the expected future reward for taking action \(a\) in state \(s\), incorporating the maximum estimated return from future states. Deep Q-Networks (DQN) and policy-gradient methods are widely used approaches in DRL, enabling efficient learning in complex and continuous domains. A general overview of the DRL training process employed is illustrated in Figure~\ref{Figure_graph_drl}.

\subsection{Deep Q Network(DQN)}

Deep Q-Networks (DQN) are a prominent DRL algorithms that extend the classical Q-learning approach by incorporating deep neural networks to approximate the action-value function. Traditional Q-learning methods become impractical for environments with large or continuous state spaces due to the need for an exhaustive Q-table. DQN overcomes this limitation by using a neural network parameterized by \(\theta\) to estimate the Q-values, as formulated in Equation \ref{dqn}.

To improve the stability of learning, DQN introduces two key techniques: experience replay and target networks. Experience replay stores past transitions \((S_t, A_t, R_{t+1}, S_{t+1})\) in a replay buffer, from which mini-batches are randomly sampled to break correlation between consecutive experiences. Target networks, on the other hand, help mitigate instability by maintaining a separate set of network parameters \(\theta^{-}\) that are periodically updated to track the main Q-network.

The training objective in DQN is to minimize the loss function given by:

\begin{equation}
L(\theta) = \mathbb{E} \left[ \left( Y_t - Q(S_t, A_t; \theta) \right)^2 \right]
\label{dqn_loss}
\end{equation}

where the target value \(Y_t\) is defined as:

\begin{equation}
Y_t = R_{t+1} + \gamma \max_{a'} Q(S_{t+1}, a'; \theta^-).
\label{dqn_target}
\end{equation}

The target value \(Y_t\) is computed using the frozen target network with parameters \(\theta^{-}\), preventing rapid fluctuations in the Q-value estimates. By combining neural networks with reinforcement learning principles, DQN enables efficient policy learning in high-dimensional state spaces, making it a foundational approach for various decision-making applications. The details about the training of DQN are presented in Algorithm \ref{alg:dqn}.

\begin{algorithm}[]
\caption{Deep Q-Network (DQN)}
\label{alg:dqn}
\begin{algorithmic}[1]

\State \textbf{Initialize} a Q-network, Key hyperparameters, such as the discount factor $\gamma$, learning rate $\alpha$, exploration rate $\epsilon$, replay buffer size, and training batch size.

\For {each timestep in the episode}
    \State Select an action $A_t$ using an $\epsilon$-greedy strategy:
    \State Execute $A_t$, observe the reward $R_t$ and the next state $s_{t+1}$.
    \State Store the transition $(s_t, a_t, r_t, s_{t+1})$ in a replay buffer.
    \State Sample a minibatch of previously stored transitions from the replay buffer.
    \State For each sampled transition, compute the \emph{target} Q-value using the target-network to estimate $\max_{a'} Q(s_{t+1}, a')$.
    \State Perform a gradient descent step to minimize the error between the current Q-values and the target Q-values for each sampled transition.
    \State Update the current state $s_t \leftarrow s_{t+1}$.
\EndFor

\State \textbf{Output:} The trained Q-network that can predict action values for any given state.

\end{algorithmic}
\end{algorithm}

\subsection{Proximal Policy Optimization (PPO)}

PPO is an advanced RL algorithm developed to improve both stability and efficiency in policy-gradient-based approaches. In traditional policy gradient methods, excessively large policy updates can negatively affect the training process, causing instability and performance degradation \cite{yuan2025_large_pol_dis}. PPO mitigates this challenge by introducing constraints on policy updates, ensuring smoother learning and consistent improvement. Specifically, PPO employs a clipped objective function designed to restrict policy changes within a predefined range, thereby stabilizing the learning trajectory and preventing overly aggressive updates. The objective function used by PPO is presented in Equation~\ref{objective_function}:

\begin{equation}
L(\theta) = \hat{\mathbb{E}}_{t}\left[\min\left(r_{t}(\theta)\hat{A}_{t}, \text{clip}\left(r_{t}(\theta), 1 - \delta, 1 + \delta\right)\hat{A}_{t}\right)\right]
\label{objective_function}
\end{equation}

In this formulation, \( r_{t}(\theta) \) denotes the ratio between the new policy and the old policy, \(\hat{A}_{t}\) indicates the estimated advantage at time \(t\), and the hyperparameter \(\delta\) controls the degree to which the policy updates are constrained. Through this approach, PPO effectively optimizes policies in complex and uncertain environments, demonstrating superior stability, performance, and applicability across various applications. The detail of training PPO is explained in Algorithm \ref{ppo_algorithm}.

\begin{algorithm}[ht]
\caption{Proximal Policy Optimization (PPO)}
\label{ppo_algorithm}
\begin{algorithmic}[1]
\State Initialize actor–critic networks and key hyperparameters ($\gamma$, $\lambda$, $\epsilon$, learning rate, epochs, minibatch size, target KL).
\For{each timestep}:
    \State Collect trajectories by running the current policy for $T$ steps.
    \State Compute returns, advantages, and normalize advantages.
    \For{each epoch}:
        \State Sample minibatches and recompute log-probabilities and values.
        \State Compute clipped policy and value losses; add entropy bonus.
        \State Update network parameters with gradient clipping.
        \State Stop early if KL divergence exceeds the target.
    \EndFor
\EndFor
\State \textbf{Output:} Trained actor (policy) and critic (value function).
\end{algorithmic}
\end{algorithm}

\subsection{Discrete Soft Actor-Critic (SAC)}
In contrast to PPO, SAC is an off-policy actor–critic algorithm that maximizes the expected return while maintaining high policy entropy. This entropy term encourages exploration and helps avoid premature convergence to deterministic strategies.
Since the heater control action is discrete (\textsc{on}/\textsc{off}), the SAC actor outputs \emph{logits} over the two possible actions, while two critic networks predict \(Q_1(s,a)\) and \(Q_2(s,a)\) for all discrete actions. The soft value target is computed as
\begin{equation}
V(s) \;=\; \sum_{a \in \mathcal{A}} \pi(a \mid s)\,\Big[\, Q_{\min}(s,a) \;-\; \alpha\, \log \pi(a \mid s) \,\Big],
\label{eq:sac_soft_value}
\end{equation}
where \(\alpha\) is the entropy temperature and \(Q_{\min}(s,a) \doteq \min\{Q_1(s,a),\,Q_2(s,a)\}\).
The critics are updated using a Polyak-averaged target network (with target parameters \(\bar\theta \leftarrow \tau\,\theta + (1-\tau)\,\bar\theta\)), and the actor parameters are optimized to minimize
\begin{equation}
J_{\pi} \;=\; \mathbb{E}_{s}\! \left[ \sum_{a \in \mathcal{A}} \pi(a \mid s)\,\Big( \alpha\, \log \pi(a \mid s) \;-\; Q_{\min}(s,a) \Big) \right].
\label{eq:sac_actor_objective}
\end{equation}
Automatic entropy tuning is applied to adapt \(\alpha\) toward the target entropy \(-\log\lvert \mathcal{A} \rvert\).

\subsection{Applications of Energy Management}

RL has been widely applied in energy management, demonstrating strong potential for optimizing decision-making in dynamic and uncertain environments. Recent works have explored both optimization-based and learning-based approaches, particularly focusing on the integration of renewable energy sources.

\subsubsection{Optimization-Based Methods for Energy Management}

Several studies have applied metaheuristic and optimization algorithms to enhance energy scheduling and cost efficiency. Li et al. proposed a two-stage microgrid scheduling model integrating electric vehicles (EVs) and employed the modified Manta Ray Foraging Optimization (MMRFO) algorithm to minimize generation, storage, and distribution costs\cite{li2024optimal}. Hassaballah et al. developed a real-time strategy for grid-connected microgrids using the Honey Badger Algorithm (HBA) for load shifting and battery scheduling\cite{hassaballah2024realtime}. Jamal et al. used the improved pelican optimization algorithm (IPOA) for nanogrid energy management with photovoltaic and battery systems, achieving superior performance\cite{jamal2024ipoa}. Amir et al. introduced an intelligent energy management system (IEMS) for EV charging stations using adaptive neuro-fuzzy control to optimize power flow between the grid, battery, and photovoltaic systems\cite{amir2024iems}.

\subsubsection{Reinforcement Learning Methods for Residential and Microgrid Systems}

Building on optimization techniques, RL-based approaches have been applied to enable adaptive and data-driven energy management. Wei et al. developed a dual iterative Q-learning algorithm for smart residential environments, optimizing battery charging and discharging cycles\cite{b28}. Kim et al. implemented an RL-based framework for smart buildings to dynamically regulate energy usage based on real-time data\cite{b29}. Ruelens et al. applied RL for electric water heater control, allowing systems to learn user and grid patterns to reduce energy waste\cite{b30}.

RL has also been applied to microgrids and storage systems. Foruzan et al. proposed an adaptive RL-based energy management system for microgrids to improve efficiency under varying demand\cite{b32}. Guan et al. used RL to manage domestic energy storage, reducing electricity costs by scheduling battery operations during off-peak hours\cite{b33}. These works highlight RL’s capacity to balance efficiency, cost, and operational flexibility.

\subsubsection{Deep Reinforcement Learning Methods for Energy Management}

To address growing system complexity, DRL has been employed to improve scalability and decision-making in energy control. Liu et al. introduced a DRL framework for household appliance scheduling, outperforming traditional rule-based methods\cite{b34}. Cao et al. proposed a DRL-based model for battery storage optimization under price uncertainty, degradation, and nonlinear efficiency, validated with UK electricity data\cite{cao2020deep}.

Advanced DRL algorithms have further improved energy scheduling. Yu et al. applied Deep Deterministic Policy Gradient (DDPG) for optimizing HVAC and energy storage systems, achieving 8.1\%–15.21\% cost savings\cite{yu2019deep}. Abedi et al. developed a real-time Q-learning-based control system for solar-integrated residential batteries, reducing monthly costs across multiple households \cite{abedi2022battery}.

In battery management, Wei et al. presented a DDPG algorithm for fast lithium-ion charging with temperature and degradation constraints\cite{wei2021deep}. Huang et al. applied PPO for solar battery capacity scheduling with built-in safety control\cite{huang2020deep}. Cheng et al. introduced a Periodic Deterministic Policy Gradient (PDPG) algorithm for multi-battery scheduling, reducing power costs by 8.79\%\cite{cheng2023reinforcement}. Paudel et al. used an MDP-based framework for large-scale storage management across 150 fast-charging stations in the PJM region\cite{paudel2023deep}.

\subsubsection{Federated and Multi-Agent Reinforcement Learning Approaches for Energy Management}

Recent studies have explored federated learning (FL) and multi-agent reinforcement learning (MARL) for scalable and privacy-preserving energy control. Li et al. proposed a federated multi-agent DRL framework with a physics-informed reward structure for multi-microgrid coordination, enabling learning without data sharing\cite{li2023federated}. Rezazadeh and Bartzoudis applied a similar FL-based method for distributed building energy management, improving efficiency and data privacy\cite{rezazadeh2022federated}.

In MARL research, Zhang et al. developed a distributed multi-agent DRL architecture for interconnected microgrids, achieving coordinated control and improved reliability\cite{zhang2023multi}. Zhou et al. introduced a Bayesian MARL approach to handle communication failures in microgrid energy management, ensuring robust operation under uncertainty\cite{zhou2021multi}.

Recent advancements in AI have demonstrated the growing use of machine learning and deep learning techniques across dairy farming operations, from animal health monitoring and milk yield estimation to precision feeding and Behaviour analysis \cite{mahato2025dairy, hall2025smart, magana2023machine, perez2025screening, neupane2024evaluating, king2024smart}. However, while these studies have significantly improved livestock management, the application of RL for energy optimization in dairy farms remains largely unexplored.

Existing RL and DRL techniques have enabled adaptive decision-making across smart grids, residential systems, and large-scale energy storage networks. However, most existing RL-based scheduling methods assume full knowledge of future prices or renewable generation, which is unrealistic in dynamic environments. Standard PPO variants also use fixed clipping or KL-divergence thresholds, often leading to unstable training when reward patterns change with tariff variations.
To address these issues, this study enhances PPO by integrating a forecasting-aware observation model with hour-of-day and month-based residual calibration, and an adaptive KL-control mechanism driven by a PID controller. The forecasting module helps the agent anticipate future changes in demand, renewable generation, and pricing, while the PID-KL controller stabilizes learning without extra tuning. Despite DRL’s success in energy systems, its use for load scheduling in dairy farming remains unexplored. Given the high energy needs and growing renewable adoption in dairy operations, applying DRL for intelligent scheduling offers a practical path toward lower costs and improved sustainability.

\section{Methodology}
\subsection{Overview of the Proposed RL Based Energy Scheduling Methods}
The proposed framework integrates with a dairy farm's energy environment using RL agents to optimize scheduling for water heating and battery operation in dairy farms. The RL agent interacts with the environment to minimize electricity costs while satisfying operational constraints. The environment provides the agent with states representing energy demand, renewable generation, and device status, and receives control actions that determine load operation. The agent is trained through repeated interactions, refining its policy to align energy usage with periods of low tariffs and high renewable energy.

\begin{figure}[ht]
\centering
\includegraphics[width=0.95\textwidth]{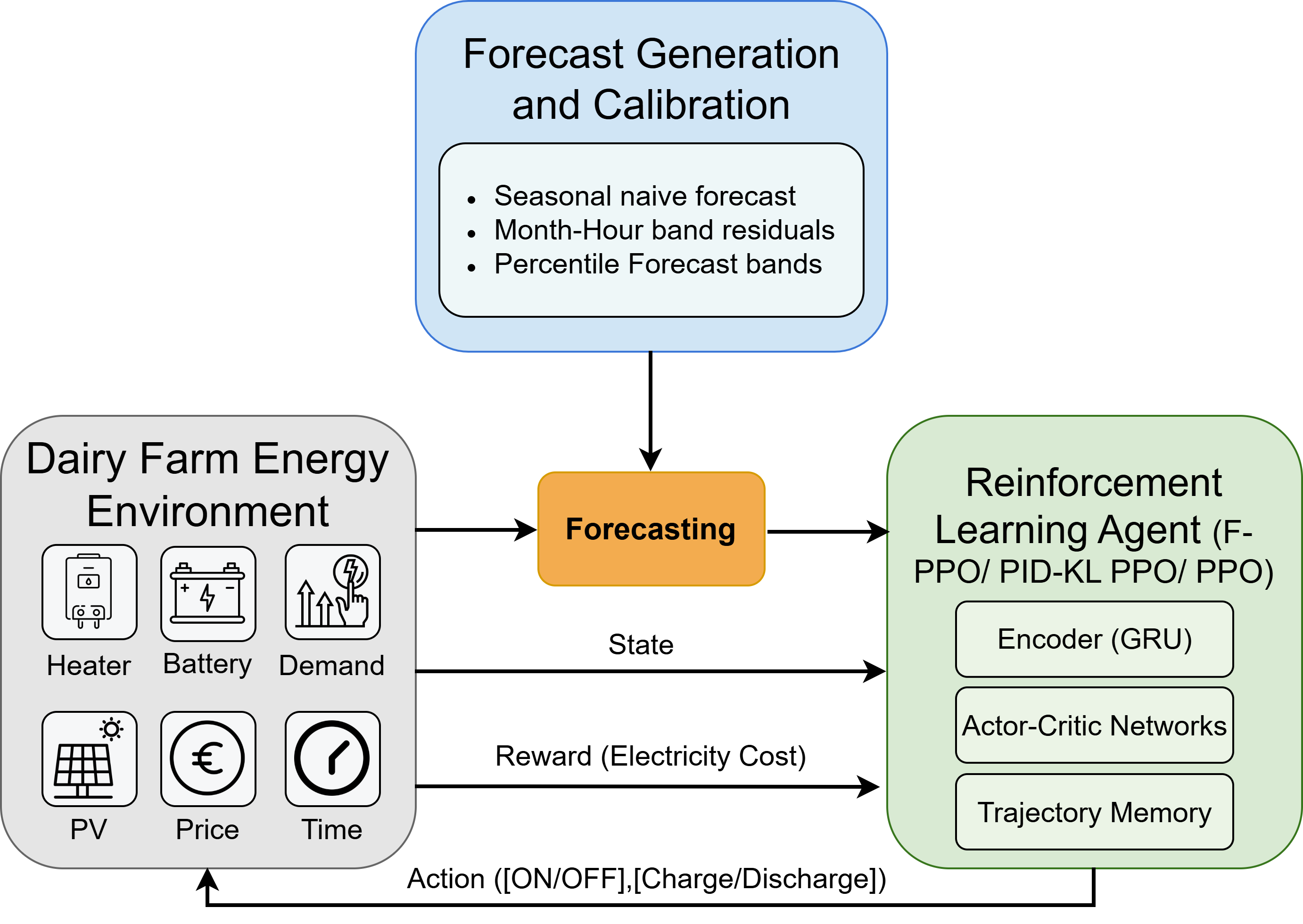}
\caption{Overview of the system environment.} \label{methodology}
\end{figure}

Figure~\ref{methodology} illustrates the overall structure of the RL-based energy management framework, where the Forecast-Aware PPO (F-PPO) and PID-KL PPO algorithms act as intelligent controllers capable of anticipating short-term variations in electricity demand, renewable generation, and electricity pricing.

\subsection{Environment Design}

The environment designed for evaluating battery and water heater scheduling consists of several essential components, including solar photovoltaic (PV) generation, battery storage(Tesla Powerwall 2.0 (13.5 kWh capacity, 5 kW charge/discharge rate)\cite{teslapowerwall}), a high-capacity water heater, the dairy farm load, and the external electricity grid. The solar PV system primarily supplies energy to meet the farm’s electricity demand or charge the battery. When renewable generation is insufficient or energy demand is high, the farm draws electricity from the external grid. A load scheduler optimizes battery and water heater scheduling decisions by analyzing real-time renewable generation, farm electricity demand, and dynamic electricity pricing to effectively minimize grid imports and operational costs.

Similarly, the environment also includes water heater scheduling. The water heater selected for this environment is the Cotswold WH 100, a high-capacity electric water heater specifically designed for dairy farms \cite{cotswold_water_heaters}. The WH 100 model features a storage capacity of 100 gallons and is rated to consume significant power (typical operation at 6 kW).  We assume that the water heater tank used in the dairy farm is thermally insulated. This is a widely adopted practice in modern dairy operations, as it allows heated water to be stored efficiently for several hours without the need for continuous energy input. The importance of tank insulation is also emphasized by the New York State Energy Research and Development Authority (NYSERDA), which highlights in its Dairy Farm Energy Best Practices Guidebook that insulating water heaters can substantially reduce heat loss and lower overall energy consumption\cite{nyserda2022dairy}. According to Dew et al., the water heater is deferrable to off-peak hours \cite{dew2021reducing}. Effective scheduling of this high-capacity water heater is crucial, as dairy farm operations require substantial hot water, and uncoordinated use significantly increases peak electricity demand and associated energy costs. The scheduling of both battery and water heater within the designed environment enables a comprehensive assessment and optimization of energy management strategies tailored to dairy farm operations.


\subsection{Data Description}

To evaluate battery scheduling operations, this research utilized dairy farm electricity consumption data in Finland, collected from a publicly accessible dataset provided by VTT Technical Research Centre of Finland \cite{fin_data}. The dataset includes hourly electricity demand for a full year, amounting to a total annual consumption of approximately 261 MW, capturing typical operational patterns and load variations of dairy farms. Renewable energy generation data were simulated using the NREL Advisor Model \cite{fin_pv}, producing hourly solar photovoltaic (PV) generation values over one year for a system with a 20 kW capacity. Electricity pricing data were retrieved from the official website of a Helsinki-based electricity supplier \cite{e_price}, featuring dynamic tariffs across three distinct price levels \cite{b_tou}. \\
For the heater scheduling analysis, we utilized a dataset from Ireland because the Finnish dataset did not provide disaggregated electricity consumption information for each equipment used in the dairy farm. The Irish dataset includes farm load profiles, PV generation data, and electricity pricing information. The load consumption data representing one year of electricity usage on a dairy farm was obtained from a study on Irish dairy farms \cite{b_hu} and contains the disaggregated information for each electrical equipment used in the farm. We selected a farm size of 200 cows, which is generally considered a large-scale dairy operation in Ireland. The PV generation data, a capacity of 20kW, was gathered from the System Advisor Model (SAM) \cite{b35}. Finally, the electricity pricing data was collected from Electric Ireland \cite{b_ir_prc}.

\subsection{Markov Decision Process(MDP) for Load scheduling}\label{state_space_battery}
The MDP framework captures sequential decision-making under uncertainty, enabling effective modeling of scheduling operations. In the MDP framework, the agent evolves its actions based on the previous actions taken and the stochastic nature of state variables such as energy demand, renewable generation, and electricity prices. By defining appropriate state variables and feasible actions, the MDP enables the formulation of an optimal policy that maximizes a specified problem objective, such as cost reduction or peak demand reduction, as well as user satisfaction.


\subsubsection{Markov Decision Process for Battery Scheduling}

\subparagraph{State Space}
The dairy farm environment for battery scheduling is represented by a state space denoted as \( S \), which encompasses all the critical information required by the algorithm for effective decision-making. The components of the state space are defined in Equation~\ref{state_space}.

\begin{equation}
\mathbf{S} = \{ hour, \text{SOC},  P_{\text{load}}, P_{\text{pv}} \}
\label{state_space}
\end{equation}

In this equation, the variable \( hour \) indicates the specific time within a 24-hour period. Incorporating the time of day in the state allows the algorithm to develop distinct strategies tailored to daily energy consumption patterns, thereby enhancing energy management efficiency. The battery's state of charge (\( SOC \)) reflects the current level of energy stored within the battery system. The SOC is discretized into 11 evenly spaced levels between 0\% and 100\%, representing approximately 10\% increments. A higher \( SOC \) allows the battery to fulfill the farm's electricity demand, minimizing reliance on grid imports. Additionally, \( P_{load} \) represents the instantaneous electricity consumption of the dairy farm. Finally, the variable \( pv_{pv} \) denotes the current solar photovoltaic (PV) generation available to the farm. PV availability directly influences scheduling decisions by determining optimal periods for energy storage or direct consumption, thus managing battery utilisation effectively.

\subparagraph{Action Space}
The actions available to the DRL agent for battery scheduling are represented by the discrete action space \( \mathbf{A} \), defined in Equation~\ref{action_space}:

\begin{equation}
\mathbf{A} = \{ Charge, Discharge, Idle \}
\label{action_space}
\end{equation}

At each decision step, the agent selects one action from this set. The \textit{Charge} action enables energy storage within the battery, typically during periods of low electricity prices or high renewable generation. Conversely, the \textit{Discharge} action allows stored energy to be utilized to satisfy farm electricity demand or to reduce dependence on grid electricity during high-price or low-renewable-generation intervals. Finally, the \textit{Idle} action is chosen when neither charging nor discharging the battery provides an immediate advantage, maintaining the battery state without energy transfer. The agent's decision-making process thus strategically balances these three actions based on real-time data to achieve optimal battery scheduling.

\subparagraph{Reward:}
The reward function $(R)$ guides the decision-making of the DRL agent by providing feedback based on the selected battery scheduling actions. Specifically, the reward is calculated considering the electricity demand of the dairy farm, the available solar power generation, and the current electricity price. Additionally, the chosen actions (charging, discharging, or remaining idle) directly influence the reward. The complete formulation of the reward function is provided in Equation~\ref{reward_battery}:

\begin{equation}
R = \begin{cases}
    -[(P_{load} + (\eta_{ch} - P_{pv})) \times E_{price}] - Penalty, & \text{if } A = Charge\\[8pt]
    -[((P_{load} - P_{pv}) - \eta_{ch}) \times E_{price}] - Penalty, & \text{if } A = Discharge\\[8pt]
    -[(P_{load} - P_{pv}) \times E_{price}], & \text{if } A = Idle
\end{cases}
\label{reward_battery}
\end{equation}

In Equation ~\ref{reward_battery}, $(P_{pv})$ denotes the instantaneous generation of solar power at time step $(t)$, and $(P_{load})$ represents the current demand for electricity from the dairy farm. The parameter $(\eta_{ch})$ corresponds to the charge or discharge rate of the battery, measured in kilowatts (kW), while $(E_{price})$ represents the electricity price at that particular timestep. The penalty term is introduced to discourage actions that negatively affect battery health or efficiency, as detailed in Equation~\ref{penalty}:

\begin{equation}
Penalty = \begin{cases}
    -15, & \text{if } SOC \geq SOC_{max} \text{ and } A = Charge\\[6pt]
    -15, & \text{if } SOC \leq SOC_{min} \text{ and } A = Discharge
\end{cases}
\label{penalty}
\end{equation}

In Equation~\ref{penalty}, $(SOC)$ denotes the battery's current state of charge, constrained between predefined minimum $(SOC_{min})$ and maximum $(SOC_{max})$ thresholds. These thresholds are set at 15\% and 85\%, respectively, to optimize battery efficiency and longevity~\cite{battery_university_808}. The penalty mechanism ensures the agent avoids inappropriate actions, such as attempting to charge when the battery is already fully charged or discharging when the state of charge is critically low. A fixed penalty value of $-15$ was applied to discourage invalid battery actions. In earlier experiments, a scaled penalty approach was tested, where the penalty was proportional to the degree of violation (e.g., how far SOC exceeded the upper or lower bounds). However, this method led the agent to adopt suboptimal behavior, where it consistently chose the least severe invalid action rather than learning to avoid such actions completely. To address this, a uniform penalty is applied to restrict the agent from taking suboptimal action.

\subsubsection{Markov Decision Process for Water Heater Scheduling}\label{state_space_heater}

In this experiment, we modeled the water heater as an MDP that balances cost efficiency with daily run-time requirements. The detail of the MDP for the water heater scheduling environment is explained below.

\paragraph{State Space}
The state space for the water heater scheduling environment is denoted as $\mathbf{S}$. The state space comprises all the information required to make optimal decisions for the water heater scheduling. The state space of the water heater scheduling is represented in Equation \ref{eq:state_space_heater}.
\begin{equation}
\mathbf{S} = \{ hour,\, E_{price},\, P_{pv},\, P_{background},\, P_{net},\, P_{device},\, run\_time \}.
\label{eq:state_space_heater}
\end{equation}

The state space represented in Equation \ref{eq:state_space_heater} includes several key elements that capture the essential factors influencing water heater operation. The \emph{hour} indicates the current time of day, such that \(0 \leq hour < 24\). The \emph{\(E_{\text{price}}\)}tracks the current electricity price, reflecting varying electricity costs at different hours. The \emph{\(P_{\text{pv}}\)} component measures the availability of renewable energy (for instance, solar power), which can offset grid consumption. The \emph{\(P_{\text{background}}\)} represents Background Power Demand, which is the electricity demand without the water heater. Meanwhile, the \emph{\(P_{\text{net}}\)} quantifies the total power consumption at each timestep, aggregating the background demand and any power consumed by the water heater. The term \emph{\(P_{\text{device}}\)} captures how much power is consumed by the water heater. It is zero when the heater is switched off and a non-zero power level when switched on. Lastly, \emph{\({run\_time}\)} denotes the number of hours for which the heater must operate during the current day to satisfy the dairy farm water requirement.

\paragraph{Action Space}
The action space of the water heater environment is represented as $\mathbf{A}$, which allows the agent to turn the water heater \emph{off} or \emph{on}. The action space is explained in Equation \ref{eq:action_space_heater}.

\begin{equation}
\mathbf{A} = \{OFF, ON\}
\label{eq:action_space_heater}
\end{equation}

In Equation \ref{eq:action_space_heater}, the agent determines the action $\mathbf{OFF}$ to switch the heater off, while  $\mathbf{ON}$ denotes switching the heater on. At each timestep, $t$, the agent selects $a_t \in \mathbf{A}$ based on the current state $s_t \in \mathbf{S}$.

\paragraph{Reward Function}
The reward function $\mathcal{R}$ encourages the agent to minimize electricity costs while meeting daily heater run-time targets within preferred time windows. Let $R_t$ be the immediate reward at timestep $t$. It is computed by combining two primary components: 
\begin{enumerate}
    \item A \textbf{cost reward}, capturing electricity expense relative to renewable availability.
    \item A \textbf{task reward}, encouraging completion of the heater's required daily operating time within desired time periods.
\end{enumerate}

We denote the cost reward as $R_{\text{cost}}(s_t,a_t)$ and the task reward as $R_{\text{task}}(s_t,a_t)$. The combined reward is given in Equation \ref{eq:combined_reward}.
\begin{equation}
R_t \;=\; \alpha \;\cdot\; R_{\text{cost}}(s_t,a_t) \;+\; \beta \;\cdot\; R_{\text{task}}(s_t,a_t)
\label{eq:combined_reward}
\end{equation}
In Equation \ref{eq:combined_reward} $\alpha$ and $\beta$ are weighting factors (e.g., $0.50$ each) to balance both reward components equally.

\subparagraph{Cost Reward.} 
Turning the heater on increases total demand; if renewable generation is insufficient, additional power must be imported from the grid, incurring cost. The formula for calculating task reward is presented in Equation \ref{eq:cost_reward}

\begin{equation}
R_{\text{cost}}(s_t,a_t) \;=\; 
\begin{cases}
    -\left[E_{price} \times 
  \bigl( (P_{background} + P_{\text{device}}) - P_{pv} \bigr) \right]   & \text{if A = ON}\\
    -(E_{price}  \times 
  \bigl( P_{background} - P_{pv} \bigr))   & \text{if A = OFF}\\
\end{cases}
\label{eq:cost_reward}
\end{equation}

In Equation \ref{eq:cost_reward}, $E_{price}$ represents the unit energy cost at the current timestamp. $P_{background}$ denotes the background power demand excluding the water heater, while $P_{device}$ refers to the electricity consumption of the water heater. The variable $A$ represents the action taken by the agent at the current timestamp, either OFF or ON. When the heater is OFF, only the background power demand is considered. $P_{pv}$ represents the renewable energy generation within the dairy farm. The primary objective of this equation is to maximize the utilisation of renewable energy while minimizing the cost of electricity imported from the grid.
\subparagraph{Task Reward.} 
The task reward ensures that the heater operates for the required duration each day, particularly within certain \emph{desired} operating periods. The variable $run\_time$ represents the remaining runtime required to turn on the water heater for the current day. After every 24 hours, $run\_time$ is checked. If it is greater than or less than zero, the agent is penalized according to Equation \ref{eq:task_reward_penalty}. If $run\_time$ is exactly zero, the agent receives a positive reward, as illustrated in Equation \ref{eq:task_reward_penalty}. The task reward is computed using Equation \ref{eq:task_reward}.

\begin{equation}
R_{\text{task}}(s_t,a_t) \;=\; 
\begin{cases}
    -(1+run\_time) + Penalty,   & \text{if A = ON}\\
    (1-run\_time - \alpha) + Penalty, & \text{if A = OFF}\\
\end{cases}
\label{eq:task_reward}
\end{equation}

Equation \ref{eq:task_reward} defines how the task reward is calculated. If the heater is ON during the desired time interval (e.g., early morning or when renewable energy generation is available), the agent receives a higher reward. Conversely, if the heater is OFF during the desired time interval, it receives a significantly higher penalty. The parameter $\alpha$ ($\alpha = 10$) is included in the equation to help the agent distinguish between beneficial and detrimental actions. The $Penalty$ is evaluated every 24 hours based on the device's $run\_time$ and is computed using Equation \ref{eq:task_reward_penalty}.

\begin{equation}
{\text{Penalty}} \;=\; 
\begin{cases}
    10,   & \text{if } \text{run\_time} = 0,\\
    -10,  & \text{if } \text{run\_time} \neq 0.
\end{cases}
\label{eq:task_reward_penalty}
\end{equation}

By balancing the objectives in Equations \ref{eq:cost_reward} and \ref{eq:task_reward}, the agent learns to switch the heater \emph{ON} strategically (to meet run-time goals and minimize cost) and \emph{OFF} when the operation is not needed or is too expensive.

\subsection{Deep Reinforcement Learning for Load Scheduling}
\subsubsection{Forecasting-Aware Proximal Policy Optimization}
The F-PPO algorithm extends the standard PPO framework by incorporating short-term forecasting signals directly into the agent’s observation space. This integration enables the policy to anticipate variations in renewable generation and electricity price, promoting proactive rather than purely reactive scheduling decisions.
In this framework, the original state representation is augmented to include planning-related variables and structured forecasting information. The complete design of the extended observation space is presented in Section \ref{sec:F_PPO_design}, which details how additional scalars and forecast matrices are constructed to provide temporal and contextual awareness to the agent.
The short-term forecasts of background power demand and renewable generation are generated using a deterministic, interpretable calibration process described in Section \ref{sec:F_PPO_calibration}. This procedure leverages a seasonal-naive model combined with hour-of-day (HOD) and month-based residual banding to represent daily and seasonal variations.
The encoded forecasts are processed using a single-layer GRU network, where the final hidden state serves as a compact representation of future dynamics. This encoded vector is concatenated with the base observation and passed through the actor and critic networks.

\subsubsection{Proportional Integral Derivative Kullback Leibler(PID-KL) PPO}
The PID-KL PPO algorithm builds upon the F-PPO framework by introducing an adaptive trust-region control mechanism. The standard PPO algorithm relies on a fixed clipping threshold or a constant KL-divergence penalty coefficient, which may not generalize effectively under dynamically changing reward magnitudes caused by fluctuating electricity tariffs. To address this limitation, an adaptive trust-region control mechanism is introduced based on a PID controller that automatically adjusts the KL penalty term, $c_{KL}$, after each update epoch.
The controller continuously monitors the deviation between the measured and target KL-divergence and updates $c_{KL}$ according to Equation \ref{eq:pid_kl}. The detail of the PID-KL update method is represented in Algorithm \ref{pid_kl_update}
\begin{equation}\label{eq:pid_kl}
c_{KL} \leftarrow \max\left(0,\, c_{KL} + K_p \cdot e + K_i \cdot \int e + K_d \cdot \Delta e \right)
\end{equation}
where $e$ represents the KL-divergence error, and $K_p$, $K_i$, and $K_d$ are proportional, integral, and derivative gains, respectively. This formulation increases the penalty when policy updates deviate excessively from the target KL (indicating instability) and reduces it when learning progress slows, thereby maintaining a balanced trust region throughout training.
By dynamically adapting the update step size, the PID-KL mechanism stabilizes training and reduces sensitivity to hyperparameter tuning. In the proposed framework, the PID-KL enhanced PPO exhibited smoother convergence behavior and greater robustness under varying electricity price conditions.

\begin{algorithm}[H]
\caption{PID-KL Controller Update in PPO}
\label{pid_kl_update}
\begin{algorithmic}[1]
\State Compute measured KL-divergence:
\[
KL_{\text{measured}} = D_{KL}\big(\pi_{\theta_{\text{old}}} \,\|\, \pi_{\theta}\big)
\]
\State Compute KL error:
\[
e = KL_{\text{measured}} - KL_{\text{target}}
\]
\State Update integral and derivative terms:
\[
I \leftarrow I + e, \quad D = e - e_{\text{prev}}
\]
\State Update adaptive KL coefficient:
\[
c_{KL} \leftarrow \max\big(0,\, c_{KL} + K_p e + K_i I + K_d D \big)
\]
\State Set $e_{\text{prev}} \leftarrow e$
\end{algorithmic}
\end{algorithm}

\subsubsection{Baseline Algorithms}
For comparative evaluation, multiple RL algorithms are implemented as baseline methods. All algorithms are trained and tested within the same environment, sharing identical state and action spaces to ensure a fair comparison. The baselines include PPO, DQN, and SAC.
For the water heater scheduling task, PPO and DQN were baseline algorithms to evaluate the effectiveness of the proposed F-PPO and PID-KL PPO frameworks. The environment is modeled as discrete actions, where the agent decides at each hourly step whether to turn the water heater ON or OFF. PPO directly learns a parameterized policy through advantage estimation and clipped surrogate objectives, while DQN adopts a value-based approach using Q-function approximation, experience replay, and an $\epsilon$-greedy exploration strategy.
To further compare the performance, the SAC algorithm is implemented for the water heater scheduling problem. The SAC model is trained on the same GRU-based forecast encoder architecture, ensuring consistent feature representation. SAC provides an off-policy comparison that benefits from replay-buffer stability and adaptive entropy regularization. The hyperparameters used for PPO, DQN, and SAC are presented in Table~\ref{tab:hyperparameters_heater}.
All baseline implementations are developed using the CleanRL \cite{huang2022cleanrl} framework to ensure reproducibility and consistent experimental procedures.

\begin{table}[ht]
\centering
\caption{Hyperparameters for Heater Scheduling}
\label{tab:hyperparameters_heater}
\begin{tabular}{|l|c|c|c|}
\hline
\textbf{Hyperparameter} & \textbf{DQN} & \textbf{PPO} & \textbf{SAC}\\ \hline
Learning rate                 & 0.00025    & 0.00025   & 0.0003    \\ \hline
Discount factor ($\gamma$)    & 0.99     & 0.99   & 0.99    \\ \hline
Initial exploration rate ($\epsilon$) & 1.0 & \textemdash & \textemdash \\ \hline
Exploration decay rate        & 0.05   & \textemdash & \textemdash \\ \hline
GRU Dropout        & \textemdash   & 0.10 - 0.15 & 0.10 \\ \hline
Clipping parameter ($\delta$) & \textemdash & 0.1   & \textemdash   \\ \hline
Batch size                & 128       & 128    & 256      \\ \hline
Replay buffer size            & 10,000   & \textemdash & 4000 \\ \hline
Total time-steps   & 1 million & 1 million  & 1 million     \\ \hline
\end{tabular}
\end{table}

For the battery scheduling problem, the PPO-based controller is evaluated against prior work by Ali et al. \cite{ali2024reinforcement}, which employed Q-learning and DQN algorithms for the same task. This comparison allows a direct assessment of the proposed PPO method against established value-based approaches under identical operational constraints, including SOC limits and dynamic electricity pricing. The hyperparameters utilized during the training of the PPO algorithm for battery scheduling are highlighted in Table~\ref{tab:hyperparameters_battery}. All hyperparameters used during training were adopted from the CleanRL implementation, which provides standardized configurations for PPO, DQN, and SAC implementations.

\begin{table}[ht]
\centering
\caption{PPO hyperparameters for Battery Scheduling}
\label{tab:hyperparameters_battery}
\begin{tabular}{|l|c|}
\hline
\textbf{Hyperparameter} & \textbf{PPO}  \\ \hline
Learning rate                 & 0.003         \\ \hline
Discount factor ($\gamma$)    & 0.89         \\ \hline
Initial exploration rate ($\epsilon$) & 1.0  \\ \hline
Exploration decay rate        & 0.0001     \\ \hline
Clipping parameter ($\delta$) & 0.2      \\ \hline
Batch size                & 64                \\ \hline
Total time-steps   & 1 million       \\ \hline
\end{tabular}
\end{table}

\subsection{Forecast-Aware Observation Design}\label{sec:F_PPO_design}
In the proposed forecast-aware PPO (F-PPO), in addition to the standard observation features from Equation \ref{eq:state_space_heater}, two planning-related variables are incorporated to enhance temporal awareness during scheduling. The first variable, \( h_{\text{left}} \), represents the remaining operational hours within the allowed daily time window, enabling the agent to estimate how much time is available to complete the required task. The second variable, referred to as \emph{slack}, quantifies the remaining scheduling flexibility and is defined as the difference between the available hours and the remaining required runtime for the current day:
\begin{equation}
\text{slack} = h_{\text{left}} - {run\_time}
\end{equation}
A positive value of \(\text{slack}\) indicates that the agent has sufficient flexibility to defer operation, whereas a negative value suggests urgency in scheduling to meet the daily heating requirement.\\
Furthermore, a \emph{forecast block} is appended to the observation vector, representing short-term forecasts of background demand and renewable generation over the next 24 hours. Depending on the configuration, the forecast block contains either two channels (median predictions for demand and PV) when \(\text{mode} = \text{one}\), or six channels (10th, 50th, and 90th percentiles for both variables) when \(\text{mode} = \text{all}\). This structured forecast input provides the agent with a probabilistic view of future dynamics, supporting proactive and cost-efficient scheduling decisions. This design allows the agent to exploit predictable variations in price and renewable supply while still respecting operational constraints. The \emph{\(h_{\text{left}}\)} and \emph{\(slack\)} scalars help the policy prioritize completion when remaining operational hours are low, thereby improving feasibility without requiring additional penalty terms in the reward.

\subsection{Forecast Generation and Calibration}\label{sec:F_PPO_calibration}
The forecasting mechanism is designed to be simple, deterministic, and interpretable, ensuring a clear separation between training and testing data. For both demand and PV generation, a seasonal-naive baseline is used as the median forecast (\(p_{50}\)), defined as
\begin{equation}
    \hat{y}^{50}_t = y_{t-24},
\end{equation}
which assumes that the hourly profile of the next day mirrors that of the same hour on the previous day. The residuals are then computed for all training data as
\begin{equation}
    r_t = y_t - \hat{y}^{50}_t.
\end{equation}
To capture both seasonal and hourly variations, residual distributions are summarized across month–hour combinations. Specifically, the 10th and 90th percentiles of residuals are computed for each \((m, h)\) pair, forming the lower (\(p_{10}\)) and upper (\(p_{90}\)) forecast bands:
\begin{equation}
    \hat{y}^{10}_t = \hat{y}^{50}_t + q_{10}(m,h), \quad
    \hat{y}^{90}_t = \hat{y}^{50}_t + q_{90}(m,h),
\end{equation}
where \(q_{10}(m,h)\) and \(q_{90}(m,h)\) represent the 10th and 90th percentile residuals for month \(m\) and hour \(h\), respectively.
All forecast series are normalized (z-scored) using the mean and standard deviation of the training months (January and July), ensuring strict data separation between training and testing. Forecasts are further clipped to the range \(\pm 5\sigma\) for numerical stability.
This residual calibration process, referred to as the \textit{HOD–Month banding}, enables the forecast representation to remain consistent across months while capturing both daily and seasonal uncertainty. Consequently, it provides the PPO agent with structured foresight on how demand and renewable generation are expected to evolve in the upcoming 24-hour window.

\section{Results and Discussion}
\subsection{Proximal Policy Optimization for Battery Scheduling}
In this experiment, we applied the PPO algorithm to effectively optimize battery scheduling operations in a dairy farming environment. The objective was to minimize dependency on the grid by maximizing renewable energy utilisation. The evaluation results demonstrate that PPO significantly reduces electricity imported from the grid, achieving a reduction of approximately 13.11\% compared to scenarios without a battery. Additionally, we compared PPO’s performance with baseline approaches, specifically Q-learning and a rule-based \cite{ali2024reinforcement}. PPO exhibited improved performance, reducing grid electricity imports by 1.62\% and 2.56\% compared to Q-learning and rule-based methods, respectively.

\begin{figure}[H]
\centering
\includegraphics[width=0.95\textwidth]{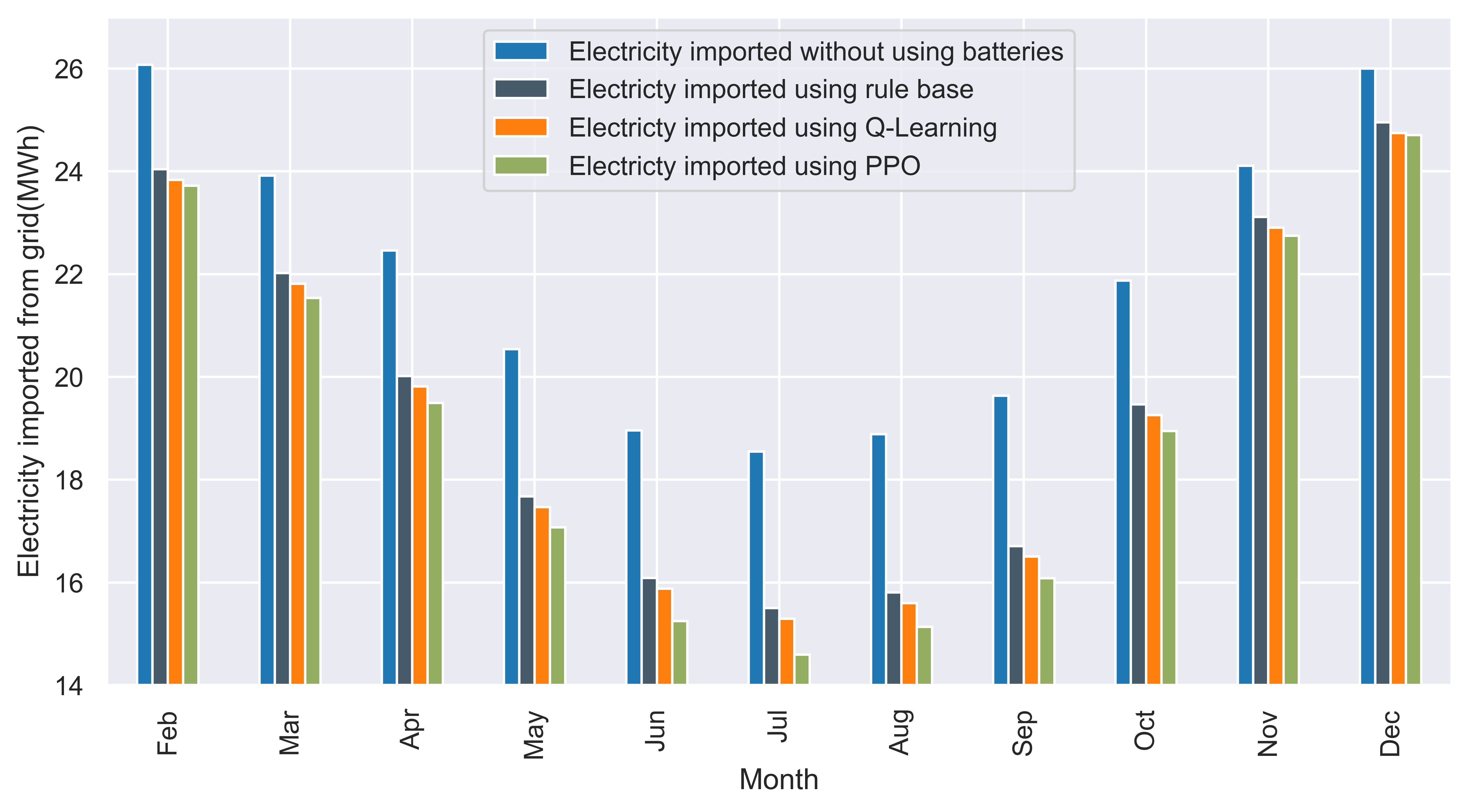}
\caption{Comparison of load imported from the grid by different algorithms.} \label{load_reduction}
\end{figure}

Figure~\ref{load_reduction} illustrates the monthly evaluation results of the algorithms from February through December. Each month displays four bars, with each bar representing the performance of one of the methodologies compared in this research. The results show seasonal variations in algorithm performance. During the summer months, PPO effectively capitalized on excessive solar energy availability, substantially reducing grid imports. Conversely, during winter months, reduced solar generation led to increased grid electricity purchases. This seasonal variation clearly demonstrates PPO’s capability to optimize battery scheduling according to renewable energy availability, effectively enhancing renewable utilisation and overall operational efficiency. The outcomes underline the potential of PPO as an effective energy management strategy within dairy farming operations.

The algorithm was trained using data from one month and evaluated on the remaining eleven months. This split was intentionally chronological rather than random to preserve the temporal structure of the data and to ensure that model performance was assessed on future, unseen conditions. Unlike conventional rule-based methods that explicitly direct agents toward predefined optimal actions, the reward function in this study was designed to allow the agent to autonomously discover effective scheduling strategies. The agent is penalized only when its actions adversely affect battery efficiency, such as attempting to charge an already full battery or discharge an empty one. This design encourages exploration within the environment, enabling the agent to develop a robust and generalizable policy derived from its own interactions. Consequently, the learned policy exhibits greater adaptability and practical relevance for real-world battery scheduling applications.

\begin{figure}[H]
\centering
\includegraphics[width=0.95\textwidth]{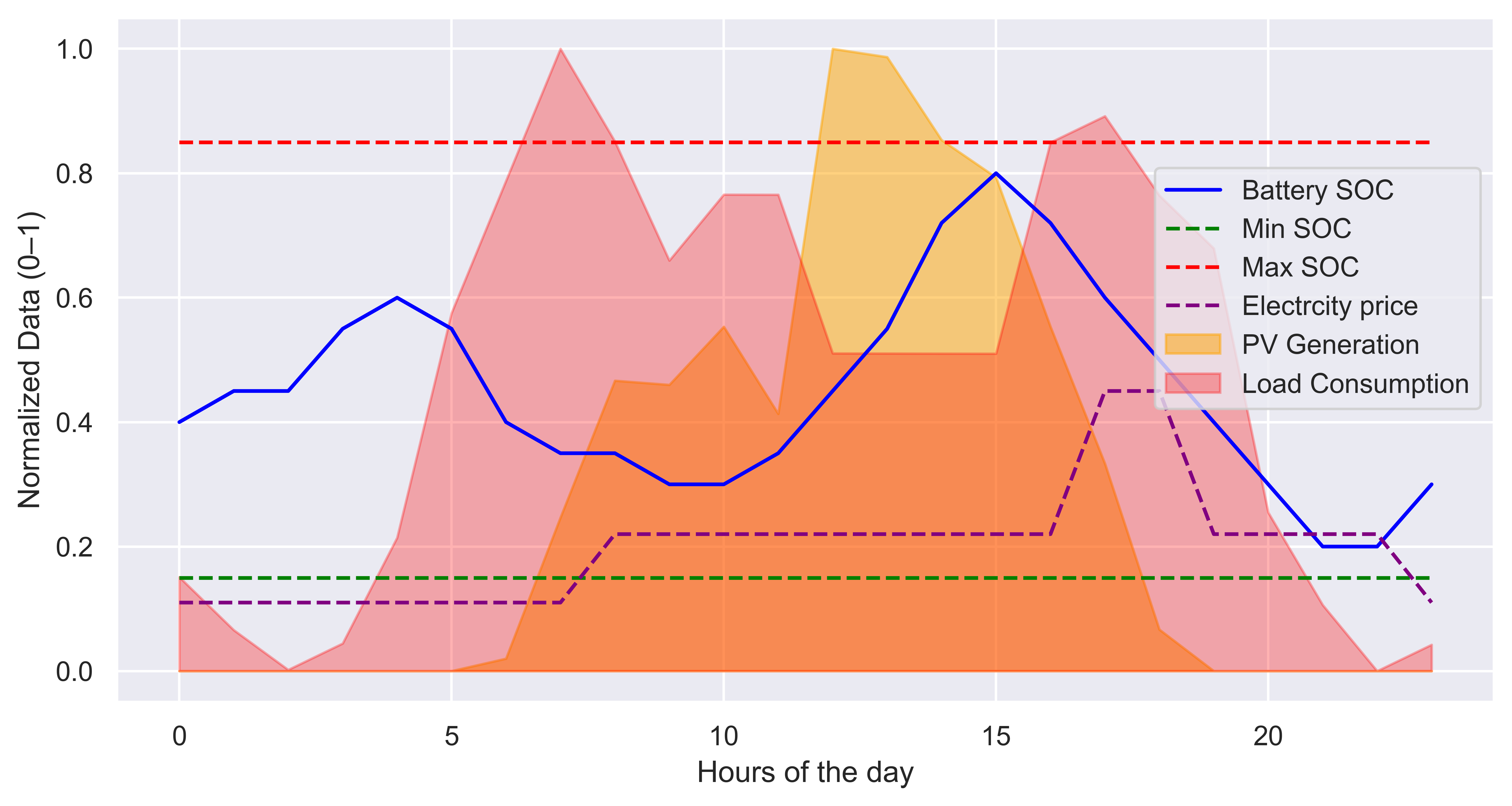}
\caption{Agent behavior for battery charging and discharging during the day.} \label{battery_control}
\end{figure}

Figure~\ref{battery_control} illustrates the daily decision-making patterns of the agent regarding battery scheduling, alongside photovoltaic (PV) generation and dynamic electricity pricing, for a randomly selected day within the year. The agent’s battery
control policy is indicated by the solid blue line, indicating the state of charge (SOC) throughout the day. The red and green dotted lines mark the maximum and minimum allowable battery charge levels, set at 85\% and 15\%, respectively, to enhance the battery's operational lifespan and maintain battery health\cite{battery_university_808}. The yellow-shaded area represents daily PV electricity generation, while the pink-shaded area shows the farm's electricity consumption pattern. Additionally, the variability in electricity pricing throughout the day is represented by the purple dotted line. The agent consistently charges the battery during periods of abundant PV generation or lower electricity prices and discharges it when renewable energy availability declines or electricity costs rise. This scheduling approach demonstrates the algorithm's capability to effectively balance renewable generation, energy demand, and cost optimization, while ensuring battery operation remains within recommended SOC boundaries.

\begin{figure}[H]
\centering
\includegraphics[width=0.95\textwidth]{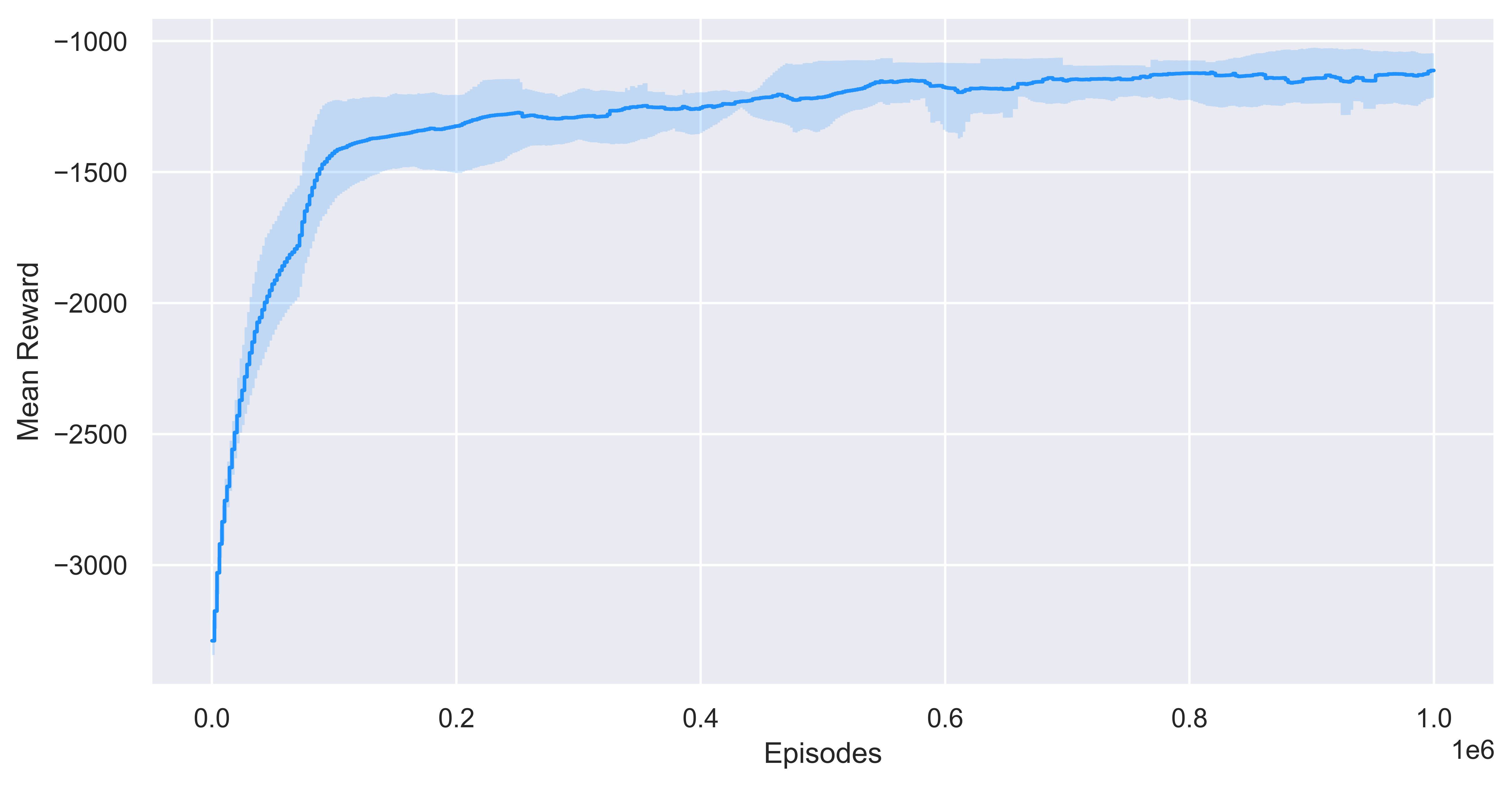}
\caption{Training reward of implemented PPO algorithm} \label{fig_reward_battery}
\end{figure}

Figure~\ref{fig_reward_battery} represents the agent’s training progression, showing average rewards and their variation throughout the learning process. The PPO algorithm’s performance was evaluated over 1 million timesteps, accumulating experiences and
adapting its policy accordingly. Initially, during the early stage of training, the rewards were low due to extensive exploration and suboptimal action selection. However, as the training progressed and the agent accumulated experience, it learned effective energy management strategies, leading to steadily improved rewards.
After approximately 0.2 million timesteps, the reward curve stabilized, indicating that the policy had effectively converged to an optimized or near-optimized solution, reflecting the agent’s ability to consistently achieve favorable battery scheduling decisions.

\begin{figure}[H]
\centering
\includegraphics[width=0.95\textwidth]{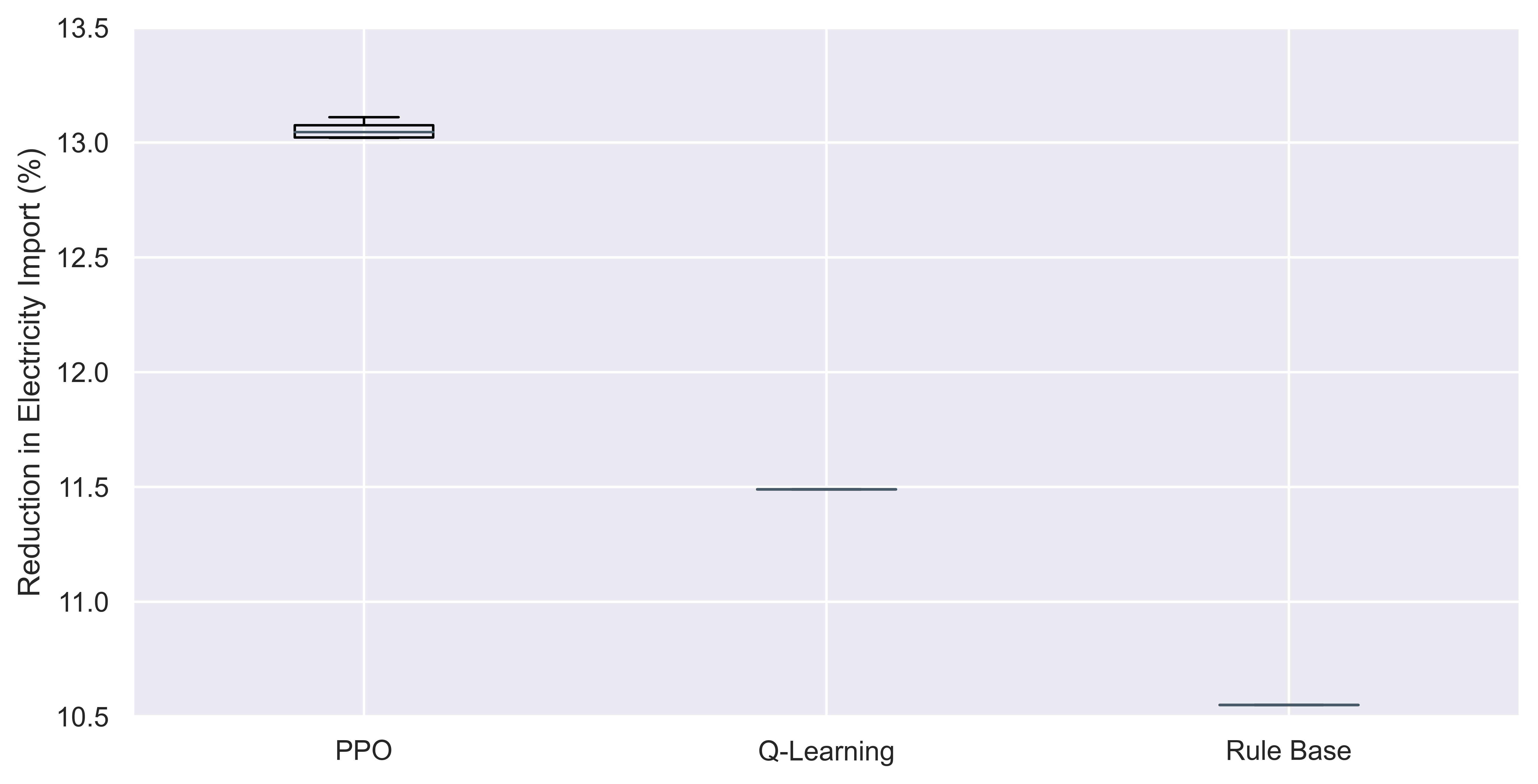}
\caption{Performance comparison of algorithms.} \label{model_comparision}
\end{figure}

Figure~\ref{model_comparision} presents a comparative analysis of load reduction percentages achieved by the PPO, Q-learning, and rule-based algorithms, represented using box plots based on results from ten independent runs over an 11-month evaluation period. The box plots illustrate the distribution and variability in performance for each algorithm. Both the rule-based and Q-learning methods exhibited relatively stable and consistent performance. In contrast, the PPO algorithm displayed greater variability in performance outcomes due to the inherently stochastic nature of its policy. PPO's policy relies on probabilistic decision-making, which can result in the selection of different actions under identical environmental conditions across various runs, thereby introducing noticeable fluctuations in performance metrics.

This variability reflects PPO's exploration capability, allowing the algorithm to discover potentially better or previously unexplored solutions compared to deterministic strategies. PPO's stochastic behavior may occasionally lead to suboptimal decisions, but more importantly, it provides greater adaptability to changing or uncertain environments. This flexibility is particularly valuable in real-world scenarios, such as dairy farm energy management, where environmental conditions and energy availability are inherently dynamic. Consequently, despite exhibiting increased variability, PPO's overall ability to achieve higher average load reductions indicates its effectiveness in adapting to complex environments compared to baseline algorithms.

\subsection{Deep Reinforcement Learning for Water Heater Scheduling}
\subsubsection{Forecasting-Aware Proximal Policy Optimization}
\begin{figure}[H]
\centering
\includegraphics[width=0.95\textwidth]{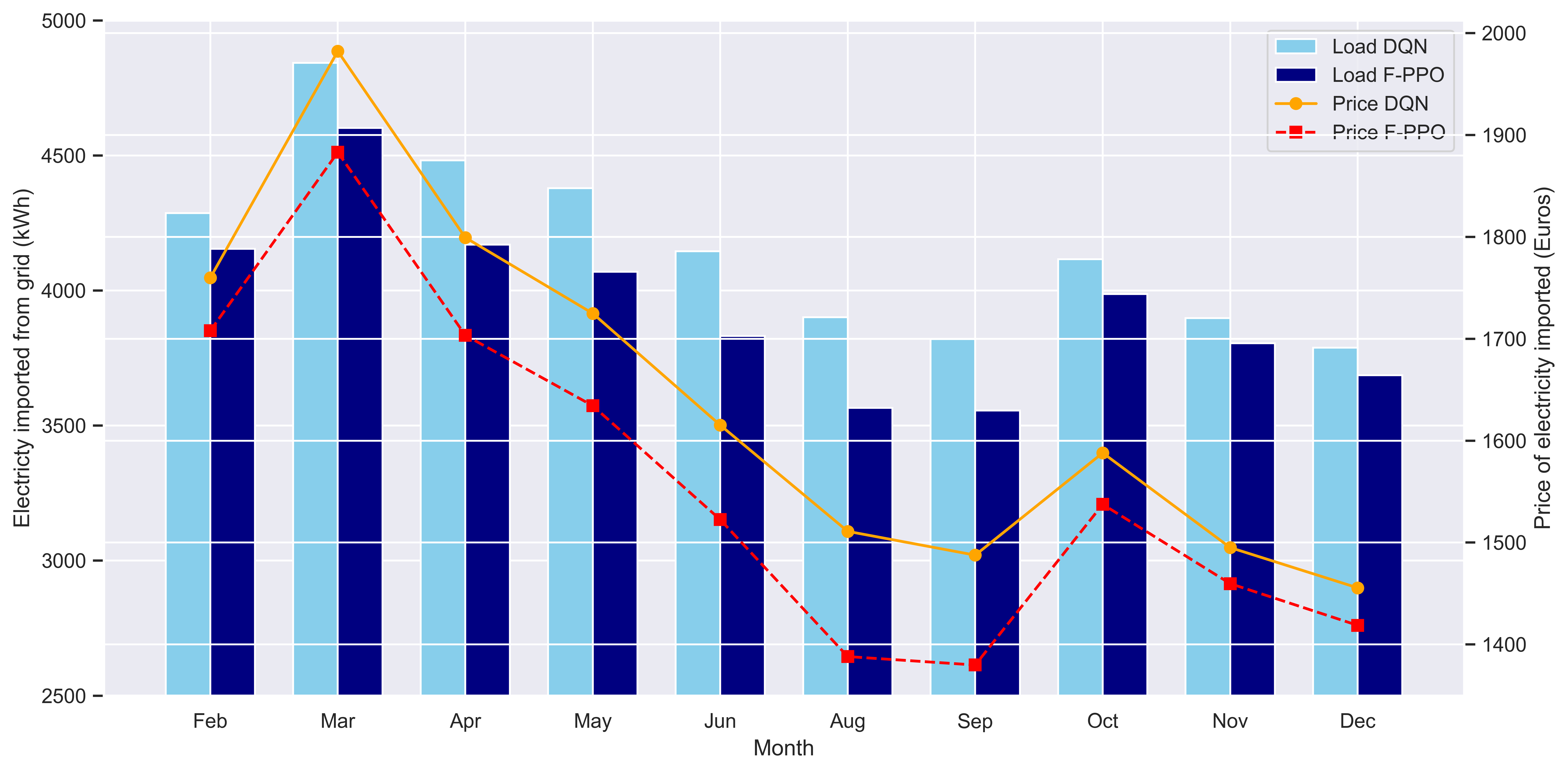}
\caption{Comparison of electricity and cost of electricity imported from the grid by using PPO and DQN.} \label{cost_load_ppo}
\end{figure}

This study investigates the scheduling of water heaters in an Irish dairy farm using F-PPO and DQN. The DQN algorithm serves as a baseline for comparison, providing insights into how the more advanced F-PPO approach enhances control decisions. Both algorithms were trained under the same conditions, utilizing the same environment, reward function, and state space information to ensure a fair and consistent performance evaluation. The hyperparameters used in the experiments are summarized in Table \ref{tab:hyperparameters_heater}.\\
The algorithm was trained using data from January and July to better generalize across different generation and consumption patterns throughout the year, and then tested on the remaining 10 months. This approach allowed for the capture of seasonal variations, ranging from low renewable energy availability in winter to higher solar generation during summer. The results demonstrate that F-PPO outperforms DQN in terms of overall cost savings, achieving a 4.76\% reduction in electricity costs. While both algorithms successfully leveraged locally generated renewable energy, PPO exhibited superior performance in shifting water heating operations to off-peak and low-tariff periods. F-PPO achieved more efficient energy management by aligning energy consumption with periods of higher renewable generation and lower electricity prices.\\
A month-by-month comparison of electricity imports and their associated costs presented in Figure \ref{cost_load_ppo} reveals that F-PPO was particularly more effective in reducing grid imports during summer, taking advantage of the greater availability of renewable energy. In winter, cost savings were less for both algorithms due to limited renewable generation. However, F-PPO maintained a competitive advantage by strategically scheduling loads during lower electricity tariff periods. These results show the ability of the implemented algorithm to schedule water heater operations during off-peak hours.

\begin{figure}[ht]
\centering
\includegraphics[width=0.95\textwidth]{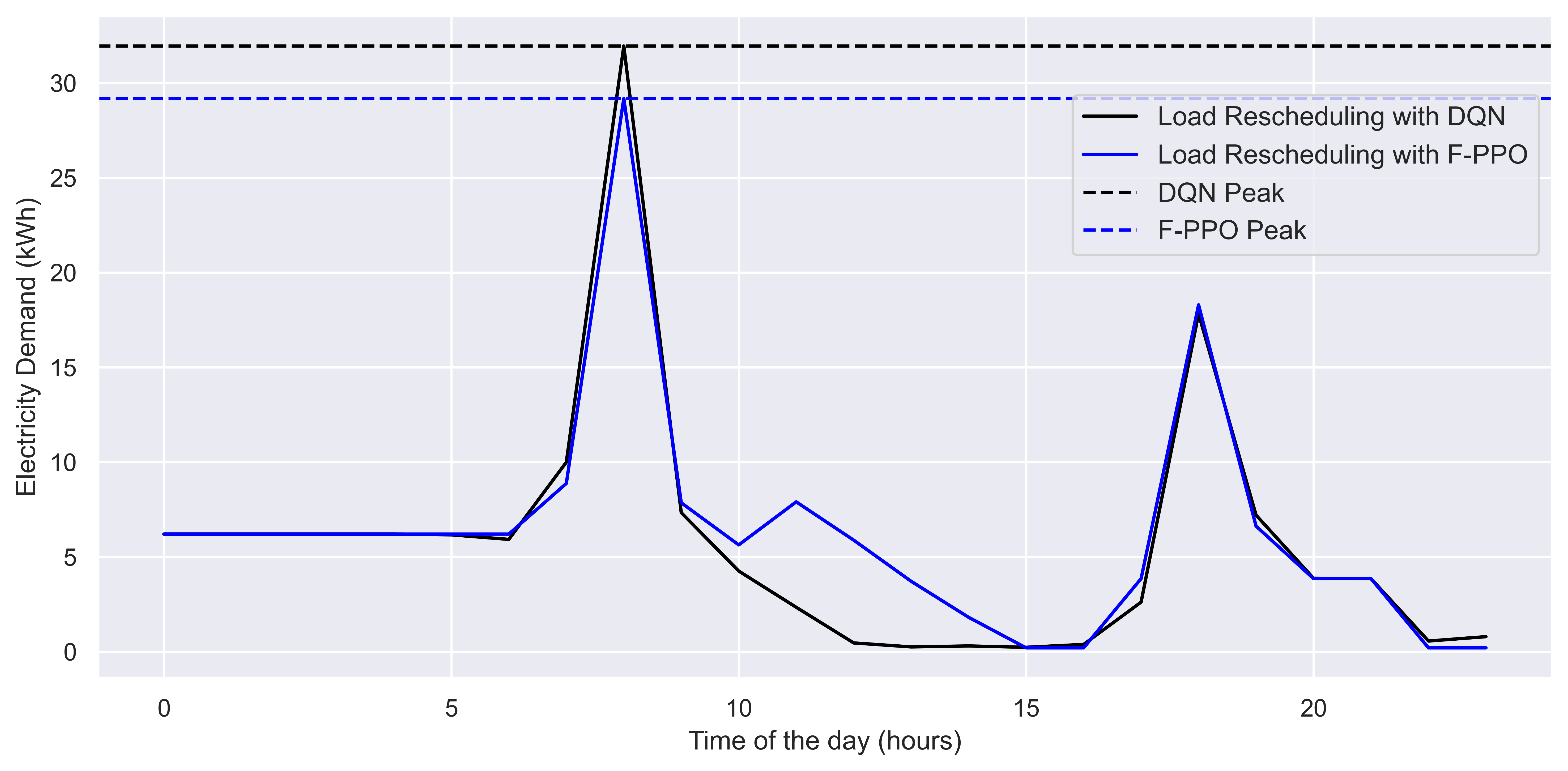}
\caption{Comparison of  average peak demand of electricity by F-PPO and DQN.} \label{peak_comparision_heater}
\end{figure}

Following the evaluation of the algorithms in terms of electricity consumption and the cost of electricity imported from the grid, we further analyzed their performance in reducing peak demand. Peak demand refers to the periods during which dairy farms experience their highest electricity consumption throughout the day. Typically, dairy farms exhibit two distinct peaks: one occurring during the morning milking session and the other during the evening milking session.\\
The results indicate that the F-PPO algorithm demonstrates a greater reduction in electricity consumption during the morning peak compared to the DQN algorithm. Specifically, the F-PPO algorithm achieves a 13.75\% reduction in average daily peak demand relative to DQN by scheduling the water heater. This finding underscores the effectiveness of F-PPO in mitigating electricity usage during high-demand periods. Figure \ref{peak_comparision_heater} illustrates the comparison of peak electricity demand in the dairy farm. The x-axis represents the hours of the day, while the y-axis denotes the corresponding electricity demand. The observed reduction in peak demand suggests that the F-PPO algorithm holds significant potential for optimizing electricity consumption in dairy farm operations.

\begin{figure}[ht]
\centering
\includegraphics[width=0.70\textwidth]{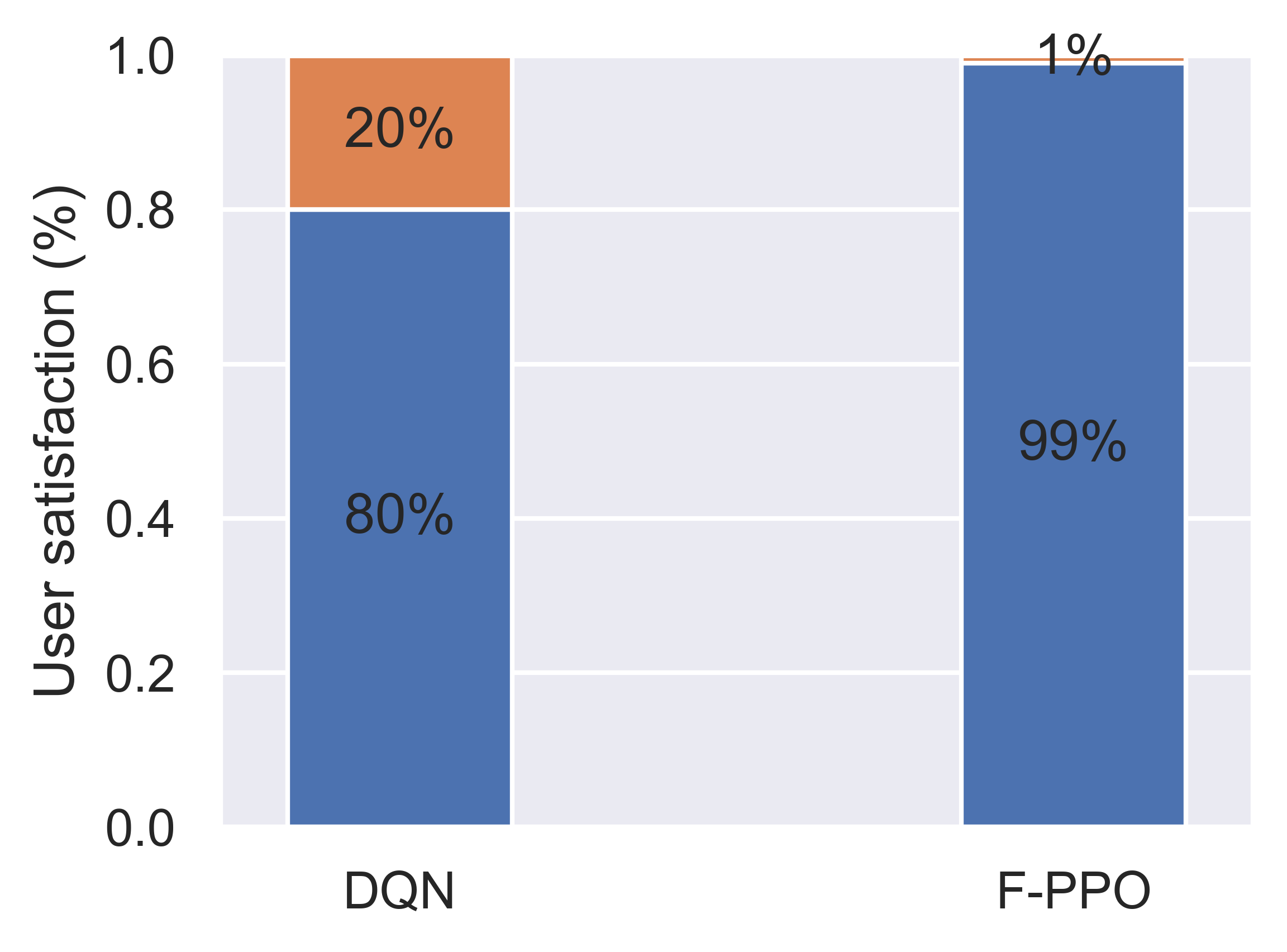}
\caption{Comparison of user satisfaction by F-PPO and DQN.} \label{satisfaction_comparision_heater}
\end{figure}

The performance of the algorithms is also assessed based on the user satisfaction rate, which is determined by whether the agent completes the required runtime for operating the water heater within the designated period or exceeds the necessary runtime. This evaluation is crucial, as failing to meet the required runtime could disrupt farm operations, while exceeding it would lead to increased electricity costs.\\
Figure \ref{satisfaction_comparision_heater} presents a comparison of user satisfaction between the two algorithms. The x-axis represents the algorithms, while the y-axis denotes the user satisfaction rate. The results indicate that the F-PPO algorithm outperforms DQN in terms of user satisfaction, achieving a high satisfaction rate of $\pm$99\%. In contrast, the DQN algorithm exhibits satisfaction rates of $\pm$80\%. These findings again highlight the reliability of F-PPO in ensuring the optimal operation of water heaters within the required time frame, thereby enhancing efficiency in dairy farm energy management.

\subsubsection{Comparative Analysis of PID-KL PPO and Forecast-Aware PPO}
\begin{figure}[ht]
\centering
\includegraphics[width=0.70\textwidth]{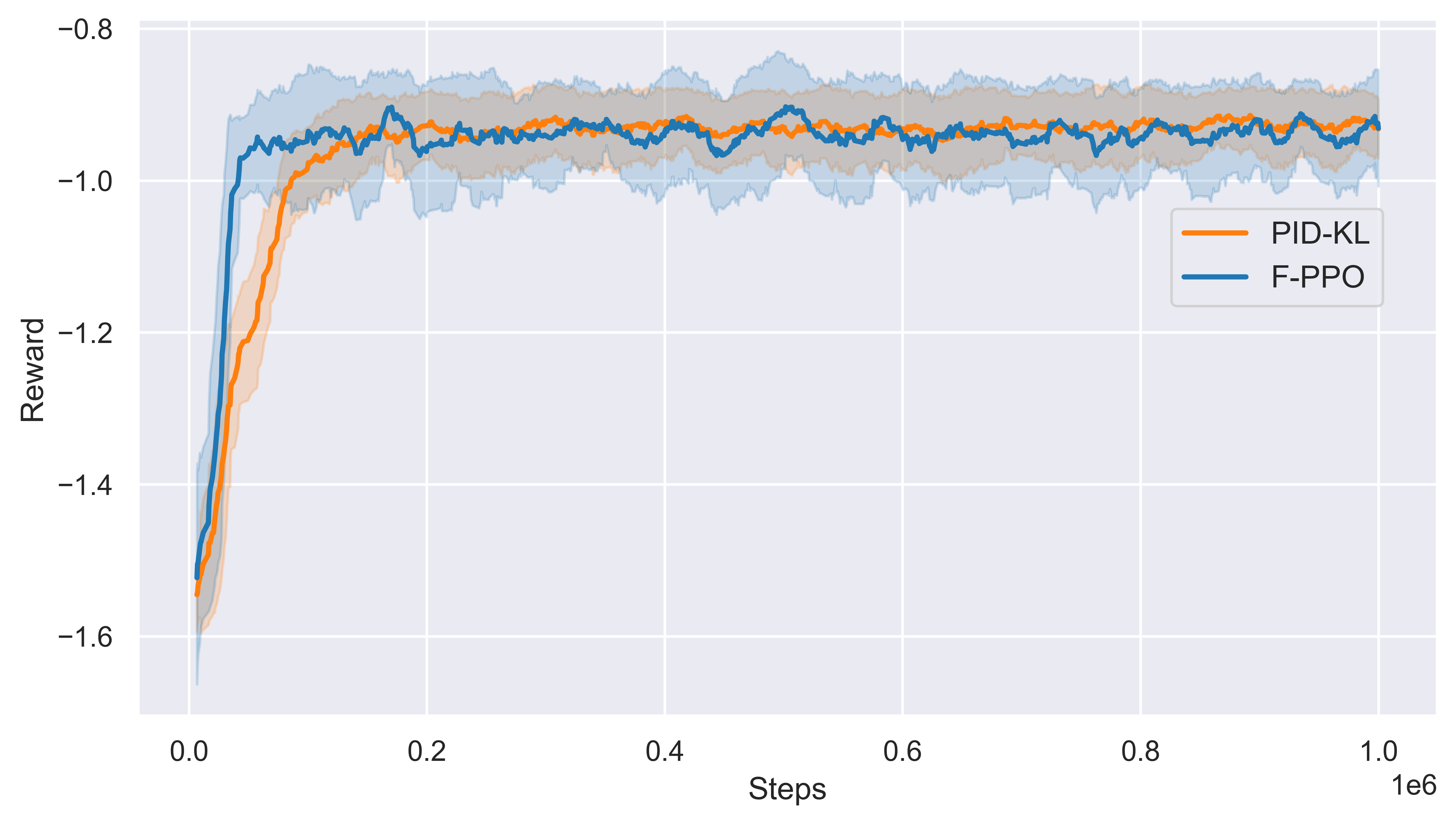}
\caption{Comparative reward convergence of PID-KL and F-PPO} \label{fig:kl_comparition}
\end{figure}

The PID-KL PPO variant extends F-PPO by incorporating an adaptive trust-region control mechanism that automatically adjusts the KL-divergence penalty during training. Instead of relying on a fixed clipping threshold, a PID controller dynamically modifies the penalty coefficient based on the deviation between measured and target KL-divergence.\\
This adaptive mechanism enhances training stability and convergence consistency, particularly under highly variable reward conditions caused by changing pricing structure. The PID-KL achieves a cost reduction comparable to F-PPO (within ±0.5\%), while demonstrating more stable policy updates and smoother learning curves across training runs. Although its final electricity savings are similar to F-PPO, the PID-KL  significantly reduces sensitivity to hyperparameter settings and prevents abrupt policy oscillations, making it more robust for long-term deployment in real-world energy scheduling systems. The Figure \ref{fig:kl_comparition} shows the comparison between the PID-KL and F-PPO. The shaded regions represent the variance across training episodes, indicating stability and consistency of policy learning. As shown, PID-KL exhibits smoother convergence with less variance compared to  F-PPO, demonstrating improved stability and controlled policy updates. In particular, PID-KL maintains a more stable reward trajectory throughout training, suggesting that its adaptive KL-divergence control effectively prevents oscillations and promotes steady policy improvement.

\subsubsection{Comparison between  PPO and Forecast-Aware PPO}
The F-PPO outperforms the standard PPO by integrating short-term forecasts of background demand and renewable generation into the observation space. This enables the agent to anticipate changes in electricity prices and renewable availability, aligning water heater operation with low-cost and high-renewable periods. Compared to the standard PPO, the proposed F-PPO achieves an electricity cost reduction of approximately 1\%, while maintaining zero overuse and under-use violations. The improvement primarily arises from the GRU-based encoder, which learns temporal dependencies within forecasted signals, allowing the agent to make proactive scheduling decisions.\\
The Hour-of-Day (HOD) and month-based residual calibration stabilizes forecast representation across unseen test months, improving generalization and reducing fluctuations in daily energy profiles. When the GRU dropout was slightly increased to 0.15 from 0.10, the policy demonstrated further cost improvement (up to 1.5\%), but the number of underused days increased, and the user satisfaction rate was 94\%, suggesting a balance between aggressive cost optimization and constraint satisfaction.

\subsubsection{Comparison with Discrete SAC}
The SAC baseline achieved competitive performance, with energy costs close to those of the PPO algorithm, but displayed higher variance across evaluation months. SAC’s entropy regularization promotes exploration and stability in continuous domains; however, in this discrete scheduling task, it required longer training durations and fine-tuning of entropy coefficients to reach convergence.\\
Despite its stability, SAC did not surpass the F-PPO variants in terms of cost reduction. Both F-PPO and PID-KL PPO maintained lower average energy costs (up to $\pm$5\% reduction compared to DQN), zero constraint violations, and consistent policy behavior. In contrast, the stochastic exploration behavior of SAC resulted in more variable scheduling decisions. Although the generated schedules remained operationally feasible, they were less consistent in aligning with low-tariff and high-renewable periods. Overall, for discrete and constraint-driven load scheduling tasks, the forecasting-enhanced PPO is more stable, interpretable, and effective in optimizing electricity usage patterns. A summary of all the experiments on water heater scheduling is presented in Table \ref{tab:algo_comparison}.

\begin{table}[H]
\centering
\small
\caption{Summary of algorithm performance on water-heater scheduling. Arrows show change vs.\ Standard PPO (↓ = lower/better, ↑ = higher/worse).}
\label{tab:algo_comparison}
\resizebox{\linewidth}{!}{%
\begin{tabular}{lcccc}
\toprule
\textbf{Algorithm} & \textbf{Total Cost (\euro)} & \textbf{Reduction vs PPO} & \textbf{Violations}  \\
\midrule
DQN         & \textit{16418} & \textit{↑ 4.10\%} & \textit{60 underuse days}        \\
Standard PPO           & 15,744 & ---       & 0               \\
Forecast-Aware PPO     & 15,635 & ↓ 0.7\%   & 0               \\
F-PPO (dropout 0.15)   & 15,582 & ↓ 1.0\%   & 17 underuse days  \\
PID-KL PPO             & 15,624 & ↓ 0.8\%   & $\pm$1 underuse days   \\
Discrete SAC           & 15,773 & ↑ 0.2\%   & 0                 \\

\bottomrule
\end{tabular}}
\end{table}

\subsubsection{Statistical Significance Analysis}
To validate whether the performance differences observed between the proposed algorithm and the comparative method DQN are statistically significant, we conducted statistical tests using the Wilcoxon signed-rank test. This non-parametric test is particularly suitable because our data distributions are not assumed to be normally distributed, and it compares paired samples effectively. We performed statistical analysis specifically for the following three key metrics:

\begin{itemize}
\item Average Daily Peak Demand Reduction
\item Electricity Imported from Grid
\item Cost of Electricity Purchased
\end{itemize}
 
The Wilcoxon signed-rank test results comparing F-PPO and DQN are summarized in Table~\ref{tab:wilcoxon_results}. The analysis reveals that the proposed PPO-based method achieves statistically significant improvements across all evaluation metrics, including electricity cost reduction, peak demand management, and grid import minimization (p-value = 0.0019). These results confirm the superior performance of the proposed algorithm in optimizing energy management, enhancing both economic efficiency and operational reliability in dairy farm water heater scheduling.

\begin{table}[H]
\centering
\small
\caption{Wilcoxon Signed-Rank Test Results of F-PPO vs DQN where ** represent the significance i.e p-value $<$ 0.05}
\label{tab:wilcoxon_results}
\resizebox{\linewidth}{!}{%
\begin{tabular}{lccc}
\toprule
\textbf{Comparison Metric} & \textbf{p-value} & \textbf{Sample size} & \textbf{Median Improvement} \\
\midrule
Average Daily Peak Demand(Month)        & 0.0019**  &  10 Months & 4.7    \\
Electricity Imported from Grid     & 0.0019**   &  10 Months  & 252.8     \\ 
Cost of Electricity Purchased     & 0.0019**    &  10 Months  & 91.6   \\ 

\end{tabular}}
\end{table}

\section{Conclusion}
This study extends the current state-of-the-art in energy management for dairy farms by developing a reinforcement learning framework that integrates forecasting, adaptive policy regulation, and real-world operational constraints. The proposed approach advances beyond traditional DRL-based scheduling by enabling proactive, stable, and constraint-compliant decision-making under dynamic electricity pricing and renewable variability.\\
A Forecasting-Aware PPO framework was introduced for water heater scheduling, allowing the agent to anticipate fluctuations in renewable generation and electricity demand through structured forecasting features. This forecasting integration enables cost-efficient scheduling while maintaining complete compliance with daily runtime requirements. Furthermore, a PID-KL adaptive controller was incorporated into PPO to dynamically regulate policy updates, improving training stability and convergence across varying electricity cost conditions. The forecasting-enhanced PPO with PID-KL achieved up to ±5\% lower electricity costs compared to DQN and approximately ±1\% improvement over standard PPO, while ensuring stable policy behavior and zero constraint violations.\\
For battery scheduling, a PPO-based approach was implemented to enable efficient energy storage management. The results showed that PPO reduced grid electricity imports by up to 13.11\% compared to scenarios without battery integration and outperformed Q-learning and rule-based strategies by 1.62\% and 2.56\%, respectively. These findings validate the proposed framework’s ability to enhance cost efficiency and flexibility in renewable energy utilisation within dairy farming operations.\\
Overall, this work demonstrates how forecasting-integrated and adaptively regulated PPO can deliver stable, intelligent, and sustainable load scheduling solutions in complex energy systems. By bridging reinforcement learning with practical energy management, the proposed framework contributes a novel, scalable direction for achieving both economic and environmental sustainability in modern agricultural settings. Future work will explore multi-agent reinforcement learning (MARL) to coordinate multiple energy-consuming devices, integrate diverse renewable sources such as wind and biogas, and employ evolutionary reinforcement learning to further improve training robustness, adaptability, and long-term optimization.

\bibliographystyle{unsrt}
\bibliography{biblography}
\end{document}